# MLMA-Net: multi-level multi-attentional learning for multi-label object detection in textile defect images

Bing Wei, *Student Member, IEEE*, Kuangrong Hao, *Member, IEEE,* Lei Gao, *Member, IEEE*

*Abstract*—For the sake of recognizing and classifying textile defects, deep learning-based methods have been proposed and achieved remarkable success in single-label textile images. However, detecting multi-label defects in a textile image remains challenging due to coexistence of multiple defects and small-size defects. To address these challenges, a multi-level, multi-attentional deep learning network (MLMA-Net) is proposed and built to: 1) increase the feature representation ability to detect small-size defects; 2) generate discriminative representation that maximizes the capability of attending the defect status, which leverages higher-resolution feature maps for multiple defects. Moreover, a multi-label object detection dataset (DHU-ML1000) in textile defect images is built to verify the performance of the proposed model. The results demonstrate that the network extracts more distinctive features and has better performance than the state-of-the-art approaches on the real-world industrial dataset.

## I. Introduction

In the modern textile industry, textile defect detection and recognition are essential to maintain product quality and processing safety. Textile defects are conventionally detected by trained workers; however, the detection precision and efficiency are low and missing rate is high because of the visual fatigue, variety of textile, and intra-class diversity of textile defects. Due to these constraints and meet the growing demand of detection efficiency, automated detection using a machine or computer is necessary for the textile industry.

Recently, single-label object detection of a textile defect image has been extensively studied (e.g., [1], [2]). Single-label textile defect detection methods can be categorized into five groups: spectral-based [3], structured-based [4], statistical-based [5], learning-based [6], [7], and model-based [8]. In particular, many methods that combine neural networks with other techniques are investigated to handle single-label textile defect detection tasks [9].

However, since the complexity of textile defects, the multi-label textile defect detection task is a more practical and general problem. This task presents two specific challenges: 1) multiple defects exist simultaneously: in the textile industry, the existence of multiple textile defects results in the intersection and interference, increasing the difficulty of accurate detection. Figure. 1(b) indicates that the multiple textile defects exist in one image. 2) small-size defects: the size of defects is small in the textile industrial processes. As shown in Figure. 1 (b), when a feature of defects is extracted by traditional methods, it is difficult to distinguish the feature correctly. Therefore, solving multi-label object detection of textile defect images is a crucial, challenging, and pressing task.

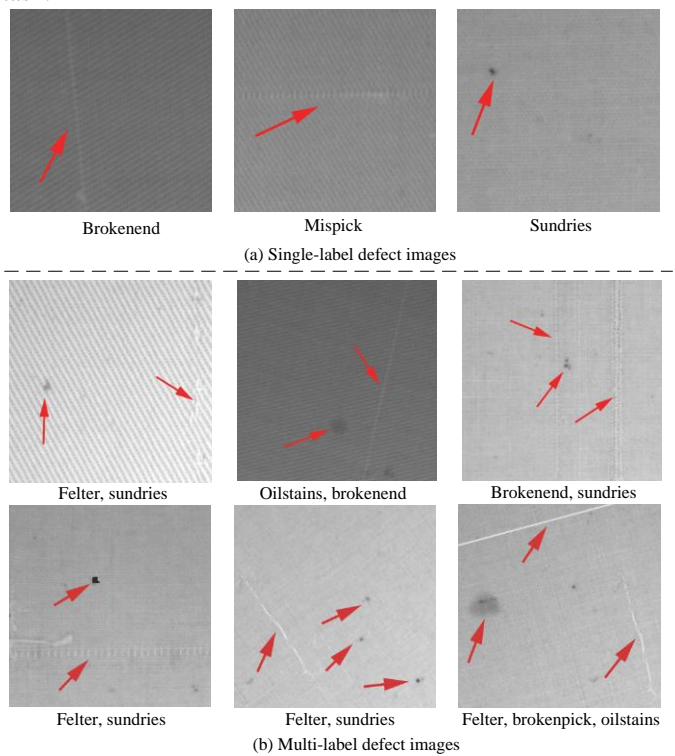

Figure 1. Examples of single-label textile defects and multi-label textile defects: (a) single-label defect images and (b) multi-label defect images.

With the rapid development of computer vision, some works attempt to address the above challenges in the industrial production. For instance, Chang et al. [10] proposed a two-step template-based correction approach for the multi-label object detection of textile defect images. Huang et al. [11] applied texture prior and low-rank representation to multi-label textile defect inspection. Also, other traditional approaches [12], [13] were used for textile defect detection, which involves multi-label object detection of textile defect images. However, these approaches need plenty of calculation to guarantee detection precision. Besides, if the dimension of the extracted feature information is low, the detection accuracies of these approaches are hard to improve.

Moreover, deep learning [14] based methods have been widely applied in computer vision tasks. As a critical deep learning method, convolutional neural networks (CNN) have demonstrated powerful discriminative ability of feature extraction in large-scale single-label image classification. Consequently, CNN is used for single-label textile defect recognition [15], textile defect detection [16], and textile pattern generation [17]. Furthermore, CNN-based defect inspection approaches have been studied for multi-label image recognition and detection. For example, Wu et al. [18] designed a weak labeled multi-label active learning method for image classification. Chen et al. [19] presented a network framework incorporating hybrid-labels in multi-label loss functions for multi-label traffic scenes classification. Wang et al. [20] proposed a multi-label object recognition network with random crop pooling for multi-label image recognition. Yao et al. [21] proposed a multi-label object detection method, which used coexistence relation network to learn the coexistence feature of multi-label objects. Li et al [22] used the visual attention mechanism to detect the multi-object defects of textile images. Although there is promising progress in multi-label image recognition/detection, the multi-label object detection of textile defect images has not been widely recognized, but it is of great significance. In addition, these existing methods were not designed to accurately detect small-size defects, as well as address the difficulty caused by the coexisting of multiple defects.

To address these challenges, this paper proposes a multi-level multi-attentional network (MLMA-Net), aiming to efficiently detect multi-label textile defects. To the best of our knowledge, this work represents the first attempt for learning multiple part modules by jointly representing multi-level and multi-attentional features. The main contributions of this paper are presented as below:

(1) We propose a MLMA-Net framework, where a multi-level module is built to obtain the refined feature information and increase the feature representation ability in detecting micro/small-size defects.

(2) A multi-attentional module is constructed to learn discriminative feature representation that maximizes the capability in attending the defect statuses.

(3) A multi-label object database for textile defect images (named as DHU-ML1000) is created for this and follow-up studies. Comprehensive experiments are conducted based on a real-world industrial dataset, and the proposed network can achieve superior detection performance over the state-of-the-art approaches.

The rest of this article is organized as follows. The relevant works are briefly summarized in Section II. Section III presents the proposed network framework. The evaluation results and discussion are provided in Section IV, followed by the conclusions and future work in Section V.

## II. RELATED WORKS

Deep learning-based methods are powerful and fast, and they can be trained in an end-to-end way. Therefore, this paper aims to use deep learning approach to detect textile defect images. In this section, we briefly introduce and review the deep neural network, multi-feature representation learning, and the related properties of attention mechanism.

### A. Deep neural network

Here, CNN is employed as the deep neural network for deep representation learning. The structure of a basic CNN includes an input layer, pooling layers, convolutional layers, a fully connected layer and an output layer, as presented in Figure. 2.

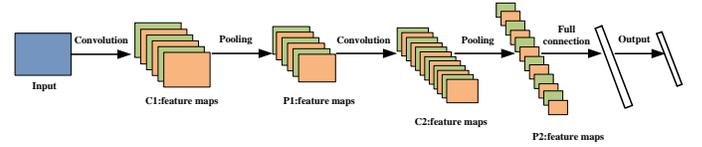

Figure. 2 The structure of a CNN.

Given an input image $X$, an array of 2-D or 3-D, we first extract the feature information of the input image by convolutional layer:

$$y_{cj}^t = f\left(\sum_{i \in P_i} k_{i,j}^t * y_{ci}^{t-1} + b_{cj}^t\right) \quad (1)$$

where $k_{i,j}^t$ denotes the convolution kernel, the mathematical symbol $*$ is the convolutional operation, $y_{cj}^{t-1}$ denotes the feature map of the $t-1$ th layer of the $j$ th channel, $b_{cj}^t$ denotes the bias, $f(\cdot)$ denotes the nonlinear activation function, $t$ is the number of layers, $y_{cj}^t$ denotes the output of feature information at the $t$ th layer of the $j$ th channel. $c$ denotes the convolutional layer parameters, which is used to distinguish the parameters of pooling layer. The pooling layer operates on each feature map, which is to progressively reduce the spatial size of the feature information:

$$y_{pj}^t = f\left(down\left(y_{ci}^{t-1}\right) \cdot w_j^t + b_{pj}^t\right) \quad (2)$$

where $down(\cdot)$ is the pooling function, $w_j^t$ is the weights, $b_{pj}^t$ is the bias of the $t$ th pooling layer, $p$ represents the pooling layer parameters, $y_{pj}^t$ represents the output of pooling operation of the $t$ th layer of the $j$ th channel. In order to enhance the fault tolerance of the network, the fully connected layer is used to perform nonlinear transformation on feature information. The output layer can indicate the predicted probability values, and is connected to the fully connected layer. Once this assignment is determined, the back propagation algorithm is used to train the whole network, and the cross entropy loss function [23] is expressed as formula (3):

$$J(W,b) = -\frac{1}{N}\sum_{i=1}^{N}\left(y^{(i)} \times \log h_{W,b}\left(X^{(i)}\right) + \left(1-y^{(i)}\right) \times \log\left(1-h_{W,b}\left(X^{(i)}\right)\right)\right) \quad (3)$$

where $y^{(i)}$ denotes its corresponding label, $h_{W,b}$ represents the predicted probability with the softmax function, $X^{(i)}$ represents the input image, $N$ represents the total number of training samples. During the training process, the low-level feature information, such as locations, edges and sizes, can be extracted by the front convolutional layer. Then, the high-level feature information is formed by combining the low-level feature information. The spatial invariance of the low-level feature information can be maintained in the high-level feature information. Also, the CNN network uses weight sharing, local connections, pooling and multiple layers to optimize its structure [24].

### B. Multi-feature representation learning

Effective multi-feature representation for image detection and recognition has attracted interest of researchers in machine learning community. A general method to deal with multi-feature is to connect all types of features into one vector. For example, the most common multiple features processing method [25] is shown in Figure. 3 (a). Figure. 3(b) illustrates the use of the multi-feature to build a feature pyramid [26]; Figure. 3(c) directly and independently uses two layers of backbone and four extra layers obtained by stride 2 convolution to construct the multiple features [27]; Figure. 3(d) constructs the multi-feature by fusing the deep and shallow layers in a top-down manner [28].

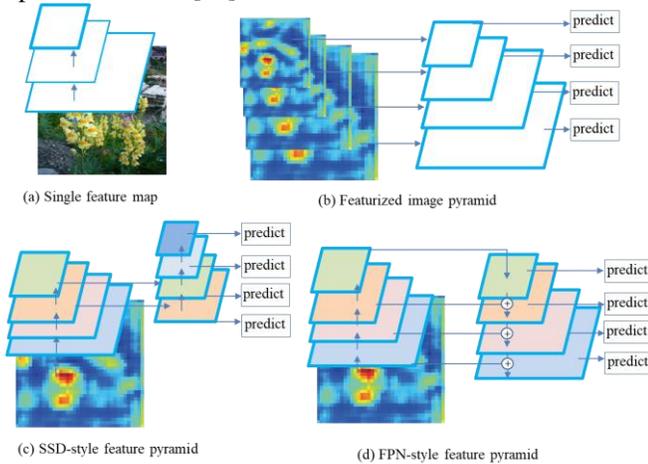

(a) Single feature map  (b) Featurized image pyramid
(c) SSD-style feature pyramid  (d) FPN-style feature pyramid

Figure. 3 Illustrations of four multiple features.

Recently, Haddad et al. [29] presented a multi-feature, sparse-based detection approach that is adaptive to the number and type of defects and features. Sindagi et al. [30] introduced a multi-feature top-bottom and bottom-top fusion scheme to merge information from multiple convolutional layers. Zhang et al. [31] proposed a multi-features aggregation framework for salient object detection. In this paper, we propose to add the multi-level module to increase the feature representation ability to detect small-size defects, which leverages higher-resolution feature maps for small textile detects.

### C. Attention mechanism

Visual attention mechanism refers that when looking at an image, only a local area of the image is focused [32]. The visual attention is characterized by exploring the surrounding environment, selecting task-related information and suppressing distraction. In one early study, the self-attention mechanism [33] is proposed to capture global dependencies of input information. Since then, visual attention has shown excellent performance in many tasks, such as text recognition, machine translation, and recommendation systems. For instance, Liang et al. [34] introduced the philosophy of attention and fixation to visual odometry. Wei et al. [35] proposed a bio-inspired integrated model that includes visual attention, visual gain and visual memory mechanisms. The visual attention mechanism allowed the model to describe and focus on the regions of interest during the multi-label textile defect image classification. Wang et al. [36] presented a deep visual attention model, which inherited the advantages of multi-level saliency information and deeply supervised network for representing hierarchy of features. Zhu et al. [37] designed an attention CoupleNet, which integrated local information and attention-related information to improve the detection performance.

In general, for each feature $v_i$ in the attention-aware feature set $V_{feat}$, the attention mechanism depends on the previous hidden state $h_{t-1}$ and generates a weight $a_i \left( a_i \in [0,1] \right)$ that represents the importance of the feature. Specifically, we have:

$$\in_{i,t} = f_{att}(v_i, h_{t-1}) \qquad (4)$$

$$\alpha_{i,t} = \frac{e^{\in_{i,t}}}{\sum_{j=1}^{m} e^{\in_{i,t}}} \qquad (5)$$

where $f_{att}$ is the multilayer perceptron network, which is the same as the model in [38]. With the weight $\alpha_{i,t}$, visual attention mechanism can obtain the context vector $z_t$:

$$z_t = \sum_{i=1}^{m} \alpha_{i,t} v_i, \qquad (6)$$

In this paper, to keep the characteristics of textile industry data, we propose the multi-attentional module to generate discriminative representation, which improves the accuracy of multi-label textile defects detection.

## III. THE PROPOSED METHOD

In this section, the proposed framework is presented in detail. The overall architecture of MLMA-Net and its three modules (i.e., feature extract network, multi-level module, and multi-attentional module) are shown in Figure. 4. Finally, the training parameters and learning procedure of MLMA-Net are described.

### A. System Overview

In this work, we define the multi-label object detection in textile defect images as the pixel-wise image-to-image task. The proposed framework is comprised of three key components: i) feature extract network; ii) multi-level module; and iii) multi-attentional module, as shown in Figure. 4.

In the framework, a batch size of preprocessed images and corresponding ground truth are inputted to the model. The basic feature information is extracted by feature extract network. These basic features are important cues for multi-label object detection in textile defect images. Then, the multi-level module is designed to extract different level high-resolution feature maps. Through the use of the multi-level module, the ability of feature representation will be enhanced in the model. Furthermore, as the characteristics of textile defects are different, and recognizing the different defects on one textile image relies on discriminative part localization. We propose the multi-attentional module to enhance the feature in the foreground regions and weaken those in the background regions. By using the multi-attentional module, the network can generate discriminative representation for the different textile defects.

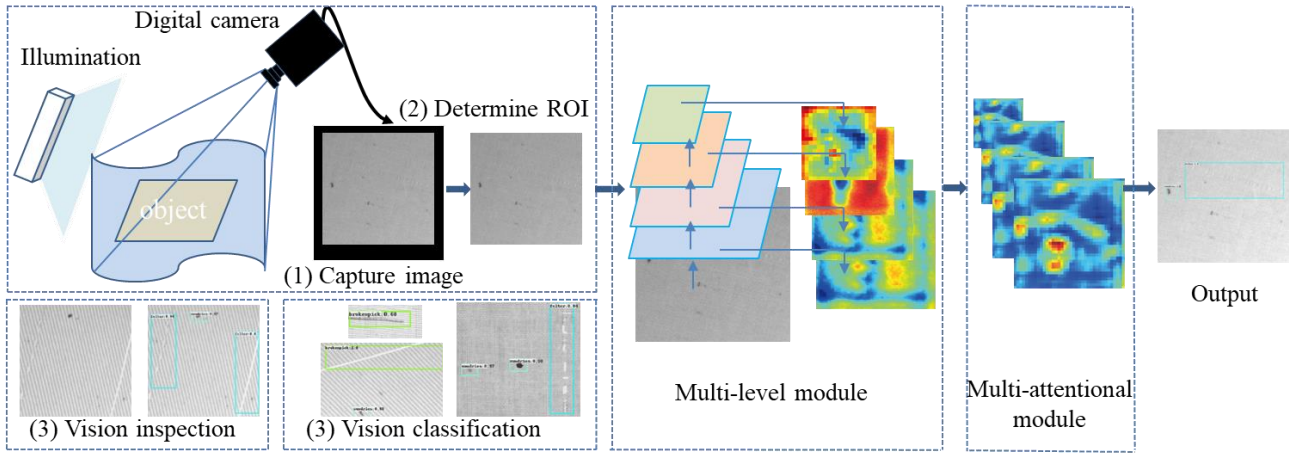

Figure. 4. The proposed MLMA-Net method. The whole network architecture consists of two key components: multi-level module and multi-attentional module. Resnet-50 network uses a pre-trained model to extract the feature maps from the input image $X$. The final output is the multi-label object detection results in textile defect images.

During the learning procedure, the network extracts the feature information of each image through the forward propagation algorithm. The feature information corresponds to the ground truth, which informs the attributes of the feature information. The feature maps of the output are used to calculate the loss during the forward propagation. Thus, the loss is minimized by the back propagation algorithm and the network can be optimized. Finally, with the help of the foreground background classifier and bounding-box regressor, the multi-label textile defect images can be recognized. The whole network architecture is shown in Figure. 5. Details of each module are presented in the next subsection.

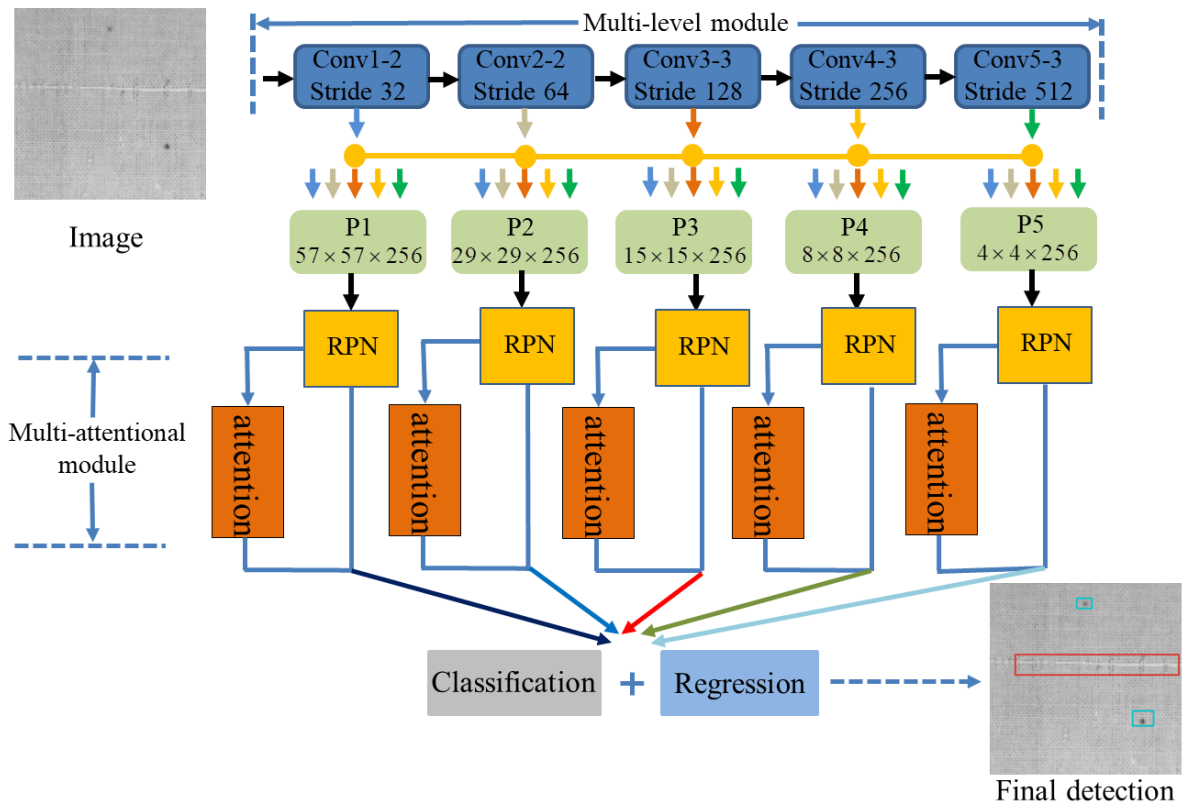

Figure. 5. The proposed network architecture of multi-label object detection for textile defect images. Note that the multi-level module processes the original

image, the multi-attentional module processes the multi-level feature maps, classification layer calculates the class by using the fully connected layer, and regression is used to obtain the final exact position and return accurate target detection box.

### B. Feature extract network (FEN)

The basic feature representation is important for the input image and has great influence on the final detection accuracy. Generally, CNN-based detectors use classification networks transferred from ImageNet classification [14] as the feature extract network. We also use the pre-trained model to initialize the parameters of the network. Given an input image $X$, the region-based features are extracted by feeding the images into the pre-trained Resnet-50 [39] network. FEN is composed of multiple processing blocks, which include convolutional layers, shortcut layers, and a single layer of average-pooling, max-pooling, and a non-linear activation layer. Each block processes the feature information through an affine transformation (convolutional layer), nonlinear activation, and feature pooling:

$$V^l = Conv(X^{l-1}) \quad (7)$$

$$O^l = f_{relu}(V^l) \quad (8)$$

$$X^l = f_{pool}(O^l) \quad (9)$$

where $Conv(\cdot)$ is a linear convolution process with two-dimension kernels and $l$ is layer index. $f_{relu}(\cdot)$ denotes the non-linear activation function. Here, the rectified linear unit (Relu) is used in the network. $f_{pool}(\cdot)$ is a subsampling operation. The output $X^l \in R^{k_l \times F_l \times T_l}$ of the $l$ th layer is a three-dimension tensor, where $k_l, F_l, T_l$ are the number of filter channels, dimensions of height and width of feature maps, respectively. In the convolutional layer, the receptive filed plays an important role in the network, which can capture feature information of the image. A large receptive filed can encode long-range relationship between pixels and leverage more context information. The context information is crucial for the subtask of localization, especially for the small objects. When FEN extracts features, early-stage feature maps are large with low-level features that describe spatial details. Late-stage feature maps are small with high-level features that are more discriminative. The low-level features are more sensitive to localization while high-level features are responsible for classification. In practice, it is known that the localization is more difficult than classification and this also indicates that early-stage feature maps are more important.

### C. Multi-level module

As mentioned in Section II, using the multi-feature representation learning, the feature representation ability can be increased in the deep learning model. In this paper, the second major component of the proposed architecture is the multi-level module, which takes the basic features of feature extract network to generate different higher-resolution feature maps. The basic features are extracted from Resnet-50 [39], which is well known for its simplicity and elegance. At the same time, Resnet-50 can yield nearly state-of-the-art results in image recognition and good generalization properties.

We propose the multi-level structure (as shown in Figure. 5), which enhances the feature representation ability to detect small defects. The implementation of this module can be divided into three steps. First, given the basic feature $X^l$, the different level high-resolution feature maps can be obtained:

$$C_2 = F_{64}(X^l) \quad (10)$$

$$C_3 = F_{128}(C_2) \quad (11)$$

$$C_4 = F_{256}(C_3) \quad (12)$$

$$C_5 = F_{512}(C_4) \quad (13)$$

For simplicity, these four features could be denoted by a feature set $F$: $F=(C_2, C_3, C_4, C_5)$, where $C_2$, $C_3$, $C_4$, and $C_5$ denote the first level, second level, third level, and fourth level features.

Second, for the sampled feature information, we use the nearest neighbor method to upsample the feature maps $C_5$, which is converted to $C_5'$.

$$C_5' = UP(C_5) \quad (14)$$

where $UP(\cdot)$ denotes the upsample operation. The upsample operation hierarchically propagates spatial information from the bottom layers to the top layers. Thus, we can adjust the channel number of $C_4$ to $C_4'$ by a $1\times1$ convolution operation.

$$C_4' = Conv_{1\times1}(C_4) \quad (15)$$

where $Conv_{1\times1}(\cdot)$ represents the $1\times1$ convolution layer. In this way, the resolutions of $C_5'$ and $C_4'$ are same, then the elements of $C_5'$ and $C_4'$ can complete the addition operation to generate the fused feature $P_5$ (similar operations are also applied to $C_3$ and $C_2$):

$$P_5' = add(C_5', C_4') = add(UP(C_5), Conv_{1\times1}(C_4)) \quad (16)$$

where $add(\cdot)$ denotes the addition operation.

Finally, the $3\times3$ convolution layers are applied to the feature information to obtain the multi-level features $P=(P_2, P_3, P_4, P_5)$. The purpose of the multi-level module is to generate different higher-resolution feature maps for detecting small textile detects. The details of the multi-level module are presented in Figure. 5. Parameters of all these layers are optimized by training on a set of labeled textile defect images to minimize the difference between prediction and ground truth. Note that the multi-level module is extremely flexible—it can be modified and replaced in various ways, such as using different layers or convolution kernel.

### D. Multi-attentional module

A number of experiments were conducted to investigate the visual attention mechanism in cognitive science [35] and artificial intelligence [36]. In this section, to address the inconsistent resolution of multi-level convolutional features and to maximize the capability of attending the defect statuses, we propose a novel multi-attentional feature representation structure—the multi-attentional module. The architecture of the multi-attentional module is illustrated in Figure. 5. Here, as one of the key innovations in our approach, the multi-attentional module can generate multiple attention maps on the multi-level convolutional features. The multi-level convolutional features with inconsistent resolution are the input of the multi-attentional module. The multi-attentional module completes the discriminative feature learning in four steps. First, given the feature map $P_i$, the batch normalization operation is used to avoid the overfitting and produce the output feature maps $f_{norm}$ as follows:

$$f_{norm} = BatchNorm2d(P_i) \quad (17)$$

where $i = (2,3,4,5)$ and $BatchNorm2d(\cdot)$ is the normalization function.

Second, we perform the sigmoid function on the $f_{norm}$. In addition, we introduce the information factor $\beta$ in sigmoid function,

$$f_{sig} = Sigmoid(\beta f_{norm}) \quad (18)$$

If $\beta = 1$, equation (18) is the sigmoid activation function, which is used to constrain the values within $[0,1]$. Equation (18) can control the output of feature information by setting the information factor $\beta$. For all experiments, it can be seen that $\beta$ ranging from 0.1 to 1 works well in practice.

Third, we re-weight the feature map $P_i$ to obtain the new feature distribution, and this can guide the model to learn the correct distribution of the foreground and background features,

$$f_{atten} = f_{sig} \otimes P_i \quad (19)$$

where $\otimes$ represents the elementwise multiplication operation between matrices. During the training procedure, the network performs supervised learning with the foreground object. The whole multi-attentional module process has two main functions. The first function is to refine the feature distribution by suppressing background features and strengthening foreground features. The second one is to help gradient flow through the layers by the multiplicative operation with information factor $\beta$.

Without equation (19), the gradient computation of back propagation is expressed in formula (20),

$$\nabla P_i = P_i' \nabla X \quad (20)$$

Using the multi-attentional module,

$$\nabla f_{atten} = f_{sig}' \nabla x \otimes P_i + f_{sig} \otimes P_i' \nabla x \quad (21)$$

Compared with equation (20), equation (21) can stabilize the training of the whole network as the multi-attentional module enables extra gradient flow in the model.

### E. Learning the proposed network

In the whole framework, the learning step of feature representation resembles ordinary network finetuning: starting from pre-trained Resnet-50 model [39] on the ImageNet dataset and training the MLMA-Net in an end-to-end way. A certain number of multi-label textile defect images are randomly selected as the training dataset. The performance of the trained model is evaluated on the testing samples.

During the training, the parameters of the new layers are randomly initialized with a Gaussian distribution $G(\mu, \sigma)$, where $\mu = 0$ and $\sigma = 0.01$. We set the initial learning rate as 0.0001. The fine-tuning of the entire framework is performed by stochastic gradient descent (SGD) with a momentum of 0.9 and weight decay of 0.0001. We conduct 20,000 iterations for fine-tuning the network. The proposed model adjusts the weights to minimize the loss function over a training set. The learning procedure is illustrated in Figure. 5.

We adopt data argument, batch-normalization, and dropout technology to limit the risk of overfitting. The deep learning library keras 2.1.3, Tensorflow 1.14.0, and relevant third-party libraries are used to develop the MLMA-Net model. The experiments are all implemented in python on a personal computer with 128GB memory and four Nvidia GeForce GTX 1080 GPUs.

## IV. EXPERIMENTAL EVALUATION AND DISCUSSION

To evaluate the performance of the MLMA-Net model, the experiments for multi-label object detection of textile defect images are conducted. In the following section, we introduce the implementation of the MLMA-Net network in detail. Furthermore, we compare its performance with other algorithms on the multi-label object detection of textile defect images (DHU-ML1000). This section includes the following four parts:

(1) The experimental setup is presented, including the dataset overview and the evaluation criteria.

(2) The new model is tested on the multi-label textile defect image dataset and the defect detection performance is evaluated through the evaluation criteria.

(3) To fully verify the performance of the proposed MLMA-Net model, the detection performance of this model is compared with that of several state-of-the-art algorithms.

(4) We discuss the experimental results and analyze the effects of some parameters on the MLMA-Net model.

### A. Datasets and evaluation

#### 4.1.1 Dataset Overview

In this study, the original textile image acquisition process is achieved by using a CCD camera. One defect-free (normal) and five defect types (sundries, brokenend, brokenpick, felter, and oilstains) are included in the multi-label textile defect images dataset (named as DHU-ML1000).

Table 1. Characteristics of textile defect image data (*n*: normal (defect-free), *be*: brokenend, *bp*: brokenpick, *f*: felter, *o*: oilstains, and *s*: sundries)

| Label Set | #Image | Label Set | #Image | Label Set | #Image |
|---|---|---|---|---|---|
| n | 50 | be + s | 155 | f + s + o | 5 |
| be | 30 | bp + s | 155 | f + bp + o | 5 |
| bp | 30 | f + s | 155 | s + bp + be | 5 |
| f | 30 | o + bp | 155 | f + s + be | 5 |
| o | 30 | s + f | 155 | s + be + o | 5 |
| s | 30 | | | Total | 1,000 |

The dataset DHU-ML1000 consists of approximately 1,000 samples, including 950 defect images and 50 defect-free textile images. The basic characteristics of the dataset DHU-ML1000 are shown in Table I. It can be seen from Table I that over 80% textile images belong to multi-label classes simultaneously. To deal with data richness and training overfitting, the dataset DHU-ML1000 is augmented by two steps: (1) textile defect segmentation and (2) data diversification. As described above, the original textile defect images are captured by the camera, with the size of each image being $1280 \times 1024$ pixels. In general, the original image may contain stain and irregular texture. Here, the first step is to crop the original image and obtain local image blocks. The segmentation result is illustrated in Figure. 6. After the segmentation operation, the size of each image is $320 \times 320$ pixels. In order to learn more invariant image features, the second step is to translate and rotate the textile defect images. The range of translation is from 0 to 50 pixels and the range of rotation is from 5º to 20º. The translation and rotation results are demonstrated in Figure. 7. During the learning and verification phase, 80% of the samples are used for training, 10% of the samples are used as the validation dataset, and the remaining samples are for prediction.

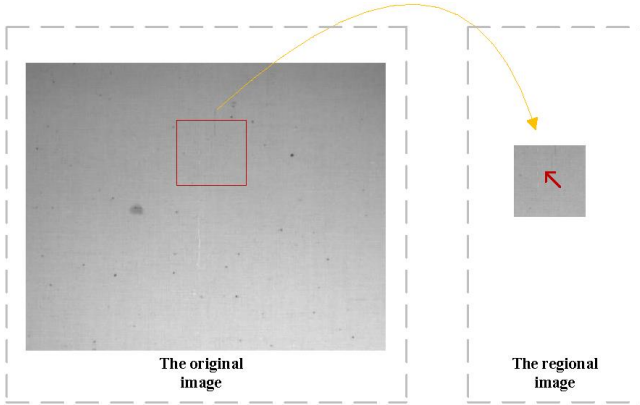

Figure. 6. Segmentation example with the original image.

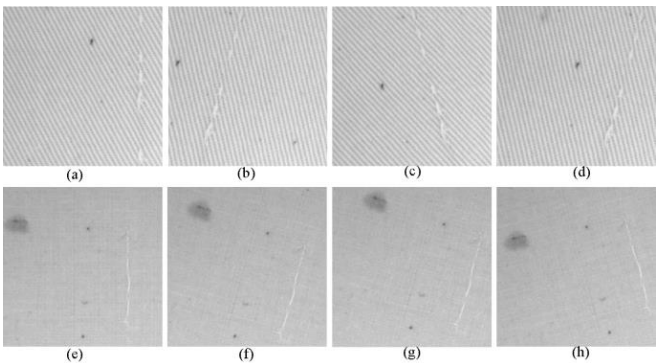

Figure. 7. Example images with translation and rotation: (a) and (e) are the original images; while (b)-(d) and (f)-(h) are the translation and rotation results.

4.1.2 Evaluation metrics

To evaluate the effectiveness of the MLMA-Net model, several multi-label indicators are selected. The evaluation metrics include Average precision (AP), mean of AP (mAP), which were used in [40]. The criteria of accuracy, precision, and recall are employed:

$$precision = \frac{TP}{TP+FP} \times 100\% \quad (22)$$

$$recall = \frac{TP}{TP+FN} \times 100\% \quad (23)$$

$$accuracy = \frac{TP+TN}{TP+FP+TN+FN} \times 100\% \quad (24)$$

where $TP$ denotes the ratio of defective samples that are detected as the defective. $FN$ denotes the ratio of defective images that are detected as defect-free. $FP$ denotes the ratio of defect-free images that are detected as defective. $TN$ denotes the ratio of defect-free images that are detected as defect-free.

B. Experiment Results

We evaluate our approach for multi-label textile defects detection and compare its detection performance with other approaches. In Figure. 8, the detection results of multi-label textile defects for some textile defect images are presented. It can be seen that the MLMA-Net model shows good performance in detecting different textile defect types. As shown in Figure. 8, the proposed network can detect the defect when there is only one defect on the textile image. Meanwhile, the proposed network can also accurately detect the defects when there are multiple defects on the textile image. Besides, for the small-size defect, such as the sundries, the proposed network can effectively detect this defect due to the multi-level module and multi-attentional module.

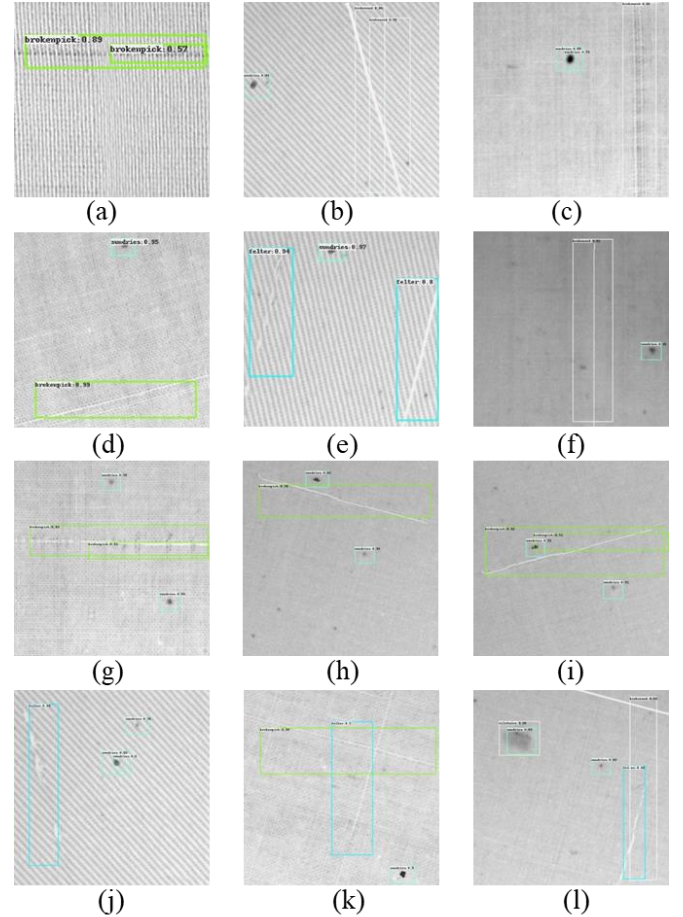

Figure. 8. The examples of multi-label textile defects detection.

The pre-class recall and AP of the proposed network are shown in Figure. 9. It is noticeable that the recall and AP of brokenpick are 98.16% and 98.36%, respectively. The recall and AP of sundries are 85.29% and 76.18%, respectively. The results also demonstrate that the recall and AP of brokenpick are better than those of the other defect types. This also implies that the detection of the small-size defect is indeed more difficult than other textile defects.

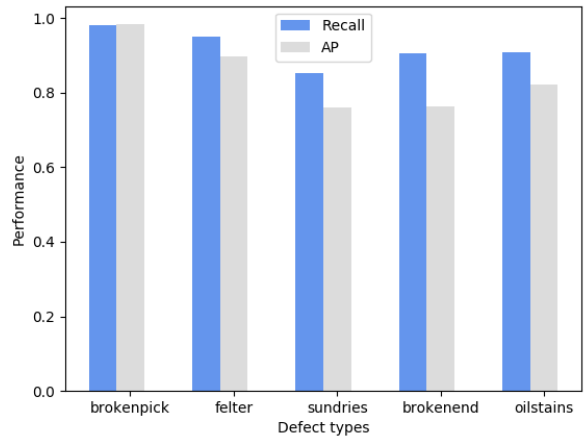

Figure. 9.The pre-class recall and AP of the MLMA-Net on the multi-label textile defect image dataset.

## C. Comparing MLMA-Net with the state-of-the-art models

The multi-label detection results of textile defect images are compared with those of the state-of-the-art methods on the dataset DHU-ML1000. Among the six selected methods, Alexnet [14] and VGG16 [41] are two popular deep learning models for image recognition. Yolo V3 [42] is a one stage object detection network, which uses the convolutional neural network structure to directly complete classification and regression tasks. RCNN [43], Faster RCNN [44], and FPN [28] are two-stage object detection networks, which first generate candidate regions (region proposals), and then classify and regress these candidate regions. Table 3 shows the multi-label detection results of textile defect images, produced by different methods. Comparing the one stage with the two-stage object detection networks, we find that the detection effectiveness of the two-stage object detection networks is overall better than the one stage one. But the detection time of the one stage network is faster than that of the two-stage ones. Compared with all the competitors, the proposed MLMA-Net network achieves the best detection performance in terms of the indicators of Accuracy, Precision, recall, and mAP. Since the proposed model is an integrated framework (with multi-level module and multi-attentional module) with more parameters to be optimized, the MLMA-Net network requires a little more time than the other ones during the testing phase.

Moreover, Table 3 shows the ablation results of the MLMA-Net without multi-level module or multi-attentional module. In the ablation study, MLMA-Net (without multi-level module) exhibits the worse performance than the MLMA-Net (without multi-attentional module). The proposed MLMA-Net network has the best comprehensive performance compared with the other methods. Compared with the performance between the MLMA-Net (without multi-level module) and MLMA-Net (without multi-attentional module), one can conclude that multi-level mechanism play a more important role than multi-attentional mechanism in the model.

Table 3. Experiment results of multi-label detection of textile defect images of different models

| Methods | Accuracy(%) | Precision(%) | recall (%) | Testing Time (s) | mAP (%) |
| --- | --- | --- | --- | --- | --- |
| Alexnet [14] | 76.41 | 57.14 | 61.90 | 0.25 | 63.85 |
| VGG16 [41] | 83.57 | 60.32 | 65.53 | 0.26 | 67.14 |
| Yolo V3 [42] | 90.48 | 66.54 | 71.63 | 0.31 | 72.68 |
| RCNN [43] | 89.72 | 63.40 | 68.81 | 0.29 | 70.63 |
| Faster RCNN [44] | 93.38 | 67.54 | 69.98 | 0.30 | 71.32 |
| FPN [28] | 94.11 | 69.22 | 72.05 | 0.61 | 75.56 |
| MLMA-Net (without multi-level module ) | 91.30 | 66.87 | 70.25 | 0.98 | 71.37 |
| MLMA-Net (without multi-attentional module ) | 95.41 | 69.87 | 73.21 | 1.20 | 76.23 |
| **MLMA-Net** | **96.69** | **70.14** | **77.78** | **1.23** | **82.13** |

Table 4. Experiment results of single-label detection of textile defect images of different models

| Methods | Accuracy (%) | Precision (%) | Recall (%) | Testing Time (s) | mAP (%) |
| --- | --- | --- | --- | --- | --- |
| Alexnet [14] | 81.25 | 71.47 | 74.83 | 0.24 | 63.65 |
| VGG16 [41] | 84.58 | 77.48 | 79.58 | 0.26 | 70.30 |
| Yolo V3 [42] | 91.11 | 84.67 | 83.33 | 0.30 | 76.42 |
| RCNN [43] | 87.28 | 80.11 | 81.91 | 0.28 | 73.36 |
| Faster RCNN [44] | 93.60 | 90.10 | 90.80 | 0.31 | 82.61 |
| FPN [28] | 95.83 | 89.36 | 91.57 | 0.63 | 82.56 |
| MLMA-Net (without multi-level module ) | 89.58 | 75.10 | 76.59 | 0.62 | 73.10 |
| MLMA-Net (without multi-attentional module ) | 93.23 | 86.46 | 89.22 | 0.63 | 79.14 |
| **MLMA-Net** | **97.50** | **91.01** | **92.18** | **0.65** | **86.07** |

It should be noted that single-label object detection of textile defect images is a special case of multi-label textile defect detection. Here, the detection performance of MLMA-Net network is also verified on the single-label textile defect detection task. The single-label textile dataset (DHU-SL1000 [45]) is adopted and expanded, which is similar to the multi-label textile dataset DHU-ML1000. Table 4 shows the experiment results of single-label detection of textile defect images of different models. As shown in Table 4, the proposed MLMA-Net framework has the best recognition performance compared with the other methods. Without multi-attentional module, MLMA-Net can obtain competitive performance with Faster RCNN. Comparing with the performance between the MLMA-Net (without multi-level module) and MLMA-Net (without multi-attentional module), we also find that multi-level mechanism is more important than multi-attentional module mechanism in single-label detection. Considering the experimental results in Tables 3 and 4, we find that the MLMA-Net framework can achieve good performance not only in multi-label textile defect detection, but also in the single-label detection.

*D. Discussion*

In the proposed MLMA-Net network, the parameters can affect the performance of the proposed model, such as batch size, learning rate, and information factor $\beta$ in multi-attentional module. In order to explore the impacts of partial parameters on detection performance, we empirically show how information factor $\beta$ in multi-attentional module affects the performance of the MLMA-Net on DHU-ML1000. As demonstrated in Figure. 10, with the increase of the information factor $\beta$, the model increasingly performs worse (higher total loss and lower mAP). We find that $\beta$ ranging from 0.1 to 1 works well in practice.

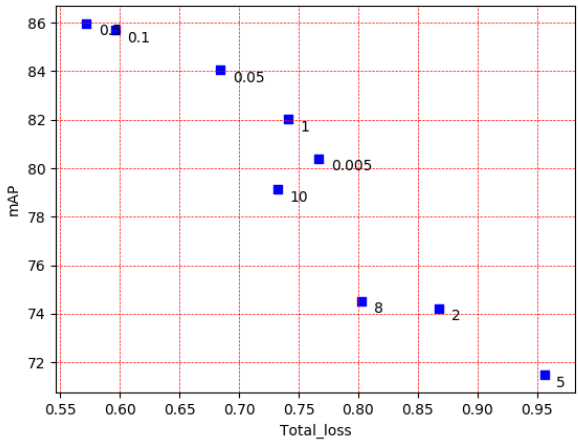

Figure. 10. Evaluation of performance with different information factor $\beta$ (represented by blue squares and associated values).

Moreover, during the learning procedure, the convergence of initial weights is an important indicator for implying whether the proposed network can converge to the stable state. Here, we track the trends of weight changes for the multi-level module and multi-attentional module. In each module, six initial weights are randomly selected and tracked. Figures. 11 and 12 present the trajectory trends of weights in the training process. It should be noted that the selected weights of each module tend to be stable finally. The experiment further confirms the stability of the proposed MLMA-Net model.

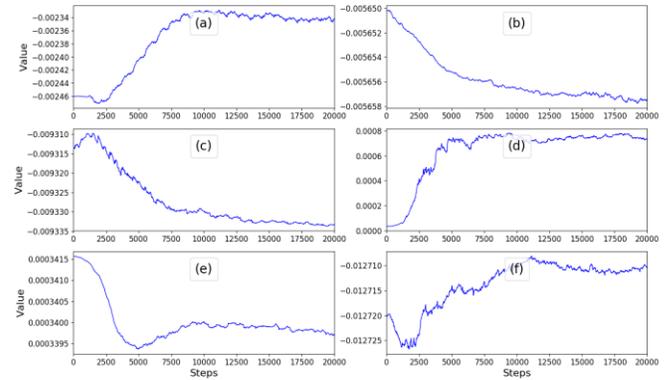

Fig. 11 Trajectory trends of weights [(a)~(f)] in the multi-level module.

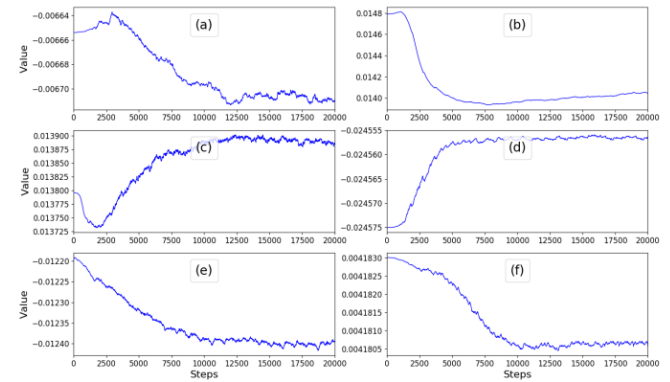

Figure. 12 Trajectory trends of weights [(a)~(f)] in the multi-attentional module.

## V. CONCLUSIONS AND FUTURE WORK

In this paper, deep learning-based method is applied to multi-label object detection of textile defect images. To overcome the shortcoming of traditional methods and existing deep learning-based methods, a novel MLMA-Net framework is proposed. The main contributions are listed as follows:

(1) We propose a novel multi-level and multi attentional convolutional feature framework, namely MLMA-Net, which is applied for the practical multi-label object detection of textile defect image in textile industry.

(2) In the model, the multi-level is designed to extract the different level high-resolution feature maps and the representation ability of features of textile defects is enhanced. The multi-attentional module is constructed to enhance the features in the foreground regions and weaken those in the background regions.

(3) For evaluating the performance of MLMA-Net, a unique multi-label object database (DHU-ML1000) of textile defect images is created for this and follow-up study. The experimental results demonstrate that the MLMA-Net framework can achieve the best performance than current state-of-the-art approaches.

Our future efforts can be devoted to applying the proposed method to solve other multi-label object detection problems such as multi-label natural scene detection and road recognition.

Besides, we would like to explore how to integrate biological mechanisms into the deep learning-based framework to expand its utility.